\documentclass{ifacconf}
\usepackage{graphicx}
\usepackage{natbib}
\usepackage{amsmath, amsfonts}
\usepackage{tikz}
\usetikzlibrary{arrows.meta,positioning,shapes,calc}

\begin{document}

\begin{frontmatter}

\title{The Driver-Blindness Phenomenon: Why Deep Sequence Models Default to Autocorrelation in Blood Glucose Forecasting\thanksref{footnoteinfo}}

\thanks[footnoteinfo]{Submitted to the IFAC World Congress 2026. This work was partially supported by The LaunchPad for Diabetes program and Capital One.}

\author[First]{Heman Shakeri}

\address[First]{School of Data Science and Center for Diabetes Technology\\University of Virginia, USA (e-mail: hs9hd@virginia.edu)}

\begin{abstract}
Deep sequence models for blood glucose forecasting consistently fail to leverage clinically informative drivers—insulin, meals, and activity—despite well-understood physiological mechanisms. We term this Driver-Blindness and formalize it via $\Delta_{\text{drivers}}$, the performance gain of multivariate models over matched univariate baselines. Across the literature, $\Delta_{\text{drivers}}$ is typically near zero. We attribute this to three interacting factors: architectural biases favoring autocorrelation (C1), data fidelity gaps that render drivers noisy and confounded (C2), and physiological heterogeneity that undermines population-level models (C3). We synthesize strategies that partially mitigate Driver-Blindness—including physiological feature encoders, causal regularization, and personalization—and recommend that future work routinely report $\Delta_{\text{drivers}}$ to prevent driver-blind models from being considered state-of-the-art.
\end{abstract}


\end{frontmatter}

\section{Introduction}

Modern deep learning architectures should, in principle, be well suited to multivariate time series forecasting. Transformers, recurrent networks, and state-space models can represent long-range dependencies, attend across channels, and integrate heterogeneous signals in high-dimensional latent spaces. In practice, however, recent analyses show that many of these models behave as sophisticated univariate forecasters: prediction performance is dominated by intra-variable autocorrelation and improves little, if at all, when exogenous covariates are added \cite{chencloser}. Large-scale benchmarks in generic time series forecasting reach a similar conclusion: carefully tuned linear models and channel-independent residual modules often match or outperform Transformer architectures, suggesting that much of the apparent gain comes from learning deep autocorrelation rather than consistently exploiting multivariate structure \citep{zeng2023transformers, das2023long}.

Blood glucose forecasting for people with Type~1 diabetes provides a particularly sharp instance of this broader phenomenon. In contrast to many application domains where the causal role of potential drivers is ambiguous, glucose dynamics are governed by well-established pharmacokinetic and pharmacodynamic mechanisms \citep{bergman1981physiologic, man2014uva}. Rapid-acting insulin has a characteristic onset, peak, and tail; meal absorption produces stereotyped increases in blood glucose over 60--120 minutes \citep{dalla2007meal}; physical activity transiently changes insulin sensitivity \citep{kemmer1992prevention}; circadian rhythms modulate insulin sensitivity over the day \citep{egi2007circadian, polle2022glycemic}; menstrual cycles affect glucose control in a subset of individuals, though with substantial heterogeneity \citep{trout2007menstrual, gamarra2023menstrual}. From a modeling perspective, insulin doses, carbohydrate intake, and activity should be highly informative drivers that enable forecasts beyond what continuous glucose monitoring (CGM) history alone can provide \citep{woldaregay2019data}.

Algorithmically, the glucose forecasting literature has followed the broader evolution of time-series modeling. Early work relied on linear autoregressive models and Kalman-filter–type estimators, often coupled to simplified compartmental physiology \citep{palerm2005hypoglycemia, turksoy2013multivariable}. Subsequent studies introduced nonlinear machine-learning predictors and hybrid schemes that encode insulin and meal information into physiologically meaningful features such as insulin-on-board (IOB), carbs-on-board (COB), and glucose rate-of-appearance curves before feeding them to neural networks \citep{bertachi2018prediction, zecchin2012neural}. More recent approaches adopt deep sequence architectures, including CNNs and LSTMs \citep{munoz2020deep, armandpour2021deep, li2020glunet}, probabilistic and quantile-based forecasters \citep{eberle2023blood}, Temporal Fusion Transformers and related attention models \citep{zhu2023edge, sergazinov2023gluformer}, and personalized or pre-trained variants that learn patient embeddings or shared representations across cohorts \citep{daniels2022multitask, deng2024pretrained}. Across these algorithmic families, reported errors at standard horizons have steadily decreased, suggesting substantial progress on the forecasting task.

Empirical results tell a different story when we ask \emph{how much} these models actually gain from driver information. We define ``Driver-Blindness'' through the metric
\begin{equation}
\Delta_{\text{drivers}} = \mathbb{E}\big[L(f_{\text{uni}}) - L(f_{\text{multi}})\big],
\label{eq:delta_drivers}
\end{equation}
where $L$ is a loss function such as RMSE, $f_{\text{uni}}$ is a model using only CGM history, and $f_{\text{multi}}$ is a model from the same architectural family that also receives exogenous inputs. Clinically meaningful multivariate models should achieve $\Delta_{\text{drivers}} \gg 0$, indicating that driver information substantially improves forecasts relative to strong univariate baselines.

Across the glucose forecasting literature, this is rarely observed. Studies spanning Gaussian processes, recurrent networks, dilated CNNs, and Transformer-style models \citep{bertachi2018prediction, munoz2020deep, li2020glunet, armandpour2021deep, eberle2023blood, sergazinov2023gluformer, zhu2023edge} consistently report modest or negligible gains over strong CGM-only baselines when insulin and meal channels are added. Borle et al.\ \citep{borle2021challenge} used Gaussian process ensembles on forty-seven individuals with detailed diabetes diaries. The multivariate model achieved a mean absolute error of $48.65$\,mg/dL compared with $52$\,mg/dL for a trivial baseline that predicts each patient's mean glucose, corresponding to $\Delta_{\text{drivers}}$ of only a few milligrams per deciliter, which the authors described as ``unexpectedly poor.'' Zhu et al.\ \citep{zhu2023edge} studied the OhioT1DM dataset with a Temporal Fusion Transformer. They found that CGM and time-of-day accounted for over ninety percent of the learned feature importance; removing all driver channels (meals, insulin, exercise) \emph{improved} RMSE by about $0.3$\,mg/dL, i.e.\ $\Delta_{\text{drivers}} < 0$. 
A synthesis of results across diverse architectures reveals that, expressed as a percentage reduction in RMSE relative to the univariate baseline, $\Delta_{\text{drivers}}$ is typically between five and ten percent at thirty minutes, and between ten and twenty percent by sixty minutes. Given the strength of physiological priors, these numbers are alarmingly small.

Driver-Blindness is not only a missed opportunity. It also reflects a deeper causal inference failure. Most real-world diabetes datasets are generated under a behavior policy, where patients actively adjust insulin and meals in response to anticipated or observed glucose levels. Insulin doses are often administered immediately before or after meals; correction boluses are given in response to hyperglycemia; exercise is scheduled in ways that correlate with these patterns. Flexible models trained purely to minimize predictive loss on such observational data can fit these behavior-induced correlations without learning the true delayed causal effects of treatments on glucose. This flexibility–causality tradeoff is now visible in hybrid and Neural ODE architectures. Zou et al.\ \citep{zou2024hybrid} showed that when hybrid neural ODE models are trained solely on forecast RMSE, the neural component can override known mechanistic constraints. 
In synthetic and real-data experiments, they observed models that inferred insulin \emph{increases} glucose (or that carbohydrates decrease it) because treatment doses almost always coincided with high glucose states in the training data. When asked to rank treatment interventions, many flexible models produced error rates statistically indistinguishable from random guessing while maintaining strong predictive accuracy. \citet{lee2024shortcomings} demonstrated a related ``forecast–control'' paradox: an LSTM forecaster achieved substantially lower RMSE than a rule-based Loop controller, yet when used to drive closed-loop insulin delivery, it produced markedly worse time-in-range. In both cases, the models were well adapted to the behavior policy that generated the data but poorly grounded in the physiology that should govern interventions.

In this paper we focus on explaining \emph{why} deep multivariate models collapse into effectively univariate, autocorrelation-dominated behavior in this setting and on what design principles can mitigate this collapse. We first present an inferential framework based on the balance between internal dynamics and external evidence, describe how gradient-based optimization naturally leads to low-arousal, driver-blind regimes when training on noisy, confounded data \citep{valle2018deep}, and then structure the resulting failure modes into three interacting challenges: architectural bias (C1), fidelity gap (C2), and personalization gap (C3). We then discuss implications for model and benchmark design, synthesize current mitigation strategies, and highlight remaining gaps and recommendations.

\section{Why Models Collapse into Deep Autocorrelation}

\subsection{Arousal, internal dynamics, and external evidence}

To understand why Driver-Blindness is such a stable outcome, it is helpful to view deep forecasting models as dynamical systems that balance internal state evolution against external evidence \citep{rao1999predictive}. A convenient formalism for this balance is variational free energy from predictive processing accounts of perception \citep{friston2010free}. We draw on this framework as a conceptual analogy rather than a formal derivation; the precise mapping to deep network dynamics remains an open question. In one common approximation, free energy can be written as
\begin{equation}
\mathcal{F} \approx \frac{1}{\alpha}\,\mathcal{E}_{\text{prior}} + \mathcal{E}_{\text{like}},
\label{eq:free_energy}
\end{equation}
where $\mathcal{E}_{\text{prior}}$ is an energy term associated with maintaining internally consistent trajectories, $\mathcal{E}_{\text{like}}$ measures misfit between predictions and sensory evidence, and $\alpha$ acts as an arousal or inverse temperature parameter that scales the influence of internal dynamics relative to external input. In glucose forecasting, the analogy is natural. The internal dynamics correspond to the strong short-term autocorrelation in CGM: at five-minute intervals, glucose tends to change slowly, and persistence alone can explain much of the variance, especially at short horizons. External evidence corresponds to insulin, carbohydrate, and activity signals that perturb this trajectory through delayed, dose-dependent mechanisms. The parameter $\alpha$ can be viewed as the effective trust the model places in drivers relative to its own extrapolation from recent glucose history.

When a deep model is trained by gradient descent on a loss function dominated by short-horizon prediction errors, several forces push it toward a low-$\alpha$ regime in which internal dynamics dominate. At fifteen to thirty minutes, the prior term associated with autocorrelation is highly predictive; many datasets and objective functions weight these horizons most heavily. Meanwhile, the likelihood term associated with drivers is noisy and inconsistent due to timing errors, missing logs, and hidden confounders such as stress, sleep, illness, and menstrual cycles that modulate responses without being explicitly observed. Finally, modern architectures with residual connections and multi-head self-attention provide low-resistance pathways for the glucose channel to dominate computations, allowing the network to achieve low loss by extrapolating trends without ever learning precise driver–response mappings. In this regime, the gradient signal that would push the model to engage with drivers is comparatively weak.

The resulting behavior resembles inattentional blindness in human perception \citep{simons1999gorillas}. The network maintains a coherent internal trajectory driven by autocorrelation and becomes functionally insensitive to exogenous signals that are not consistently predictive. Once parameters move into such a low-arousal attractor, where the prior term in \eqref{eq:free_energy} yields low loss and the marginal benefit of incorporating noisy drivers is small or negative, it is difficult for training to escape. This manifests at the level of \(\Delta_{\text{drivers}}\): even complex multivariate architectures behave as ``deep autocorrelation models,'' with $\Delta_{\text{drivers}}$ close to zero or negative.

A second useful concept is that an external signal must be syntactically and temporally aligned with the patterns a system expects in order to be integrated. We borrow the term \emph{conspecificity} from biological sensory processing \citep{bjoring2021zebra, le2025zebra} to describe this necessary alignment. Inputs that are scrambled, irregularly sampled, or misaligned relative to internal timescales are treated as noise and suppressed. In glucose forecasting, driver channels are often non-conspecific with the learned internal model. Bolus and meal logs are sparse impulses measured on a fifteen-minute grid or worse; their true physiological impact unfolds over hours. Their timing is noisy; their magnitudes are estimated with substantial error; their effects are heavily modulated by unobserved factors. To a model trained to extrapolate smooth five-minute CGM sequences, such inputs appear as erratic, low signal-to-noise perturbations. In the language of the arousal–conspecificity framework, Driver-Blindness corresponds to a regime of low effective arousal and low conspecificity, in which internal dynamics dominate and drivers are ignored.

\begin{figure*}[h]
    \centering
    \begin{tikzpicture}[
        >=Latex,
        axis/.style={->,thick,gray!60},
        region/.style={rounded corners=3pt,fill=gray!8},
        labelnode/.style={font=\small},
        title/.style={font=\small\bfseries,align=center},
        desc/.style={font=\footnotesize,align=center,text width=2.8cm},
        scale=1.0
    ]
        \draw[axis] (0,0) -- (7.5,0) node[below left,black,font=\small] {Effective arousal $\alpha$ (Prior $\to$ Likelihood)};
        \draw[axis] (0,0) -- (0,6) node[above left,black, rotate = 90,font=\small,align=center] {Conspecificity\\of drivers};


        \fill[region,draw=gray!30] (0.3,0.3) rectangle (3.6,2.8);
        \fill[region,draw=gray!30] (3.8,0.3) rectangle (7.2,2.8);
        \fill[region,draw=gray!30] (0.3,3.0) rectangle (3.6,5.5);
        \fill[region,draw=gray!30,fill=blue!5] (3.8,3.0) rectangle (7.2,5.5);

        \node[title] at (2.0,2.3) {Driver-Blind};
        \node[desc] at (2.0,1.5) {Low $\alpha$,\\low conspecificity};
        \node[desc,font=\footnotesize\itshape] at (2.0,0.7) {(deep autocorrelation)};

        \node[title] at (5.5,2.3) {Unstable /};
        \node[title] at (5.5,1.9) {noise-reactive};
        \node[desc] at (5.5,1.0) {High $\alpha$,\\low conspecificity};

        \node[title] at (2.0,5.0) {Under-utilized};
        \node[title] at (2.0,4.6) {drivers};
        \node[desc] at (2.0,3.7) {Low $\alpha$,\\high conspecificity};

        \node[title] at (5.5,5.0) {\textbf{Driver-engaged}};
        \node[title] at (5.5,4.6) {regime (desired)};
        \node[desc] at (5.5,3.7) {High $\alpha$,\\high conspecificity};

        \draw[->,very thick,blue!70,line width=1.2pt] 
            (5.2,3.5) .. controls (4.2,2.8) and (3.0,2.3) .. (2.2,1.8);
        
        \draw[->,very thick,red!70,line width=1.2pt] 
            (6.2,4.5) .. controls (6.8,3.5) and (6.5,2.5) .. (5.8,1.8);

        \draw[->,very thick,purple!60,line width=1.2pt,dashed] 
            (5.8,5.2) .. controls (4.8,4.5) and (3.5,3.5) .. 
            (2.4,2.5) .. controls (1.9,2.0) .. (1.2,1.5);
        
        \node[labelnode,purple!60,font=\footnotesize\bfseries,above right] at (5.5,5.3) 
            {training ideal};
        \node[labelnode,purple!60,font=\footnotesize\bfseries,below left] at (1.7,1.3) 
            {typical outcome};

        \node[draw,rectangle,align=left,font=\footnotesize,fill=white,inner sep=4pt,anchor=north west] at (7.5,5) {
            \textbf{Forces:}\\[2pt]
            \textcolor{blue!70}{\rule{1em}{0.8pt}} \textbf{C1:} Architectural\\
            \hspace{1.2em} shortcuts favor\\
            \hspace{1.2em} autocorrelation\\[3pt]
            \textcolor{red!70}{\rule{1em}{0.8pt}} \textbf{C2:} Fidelity gaps\\
            \hspace{1.2em} make drivers\\
            \hspace{1.2em} non-conspecific\\[3pt]
            \textcolor{purple!60}{\rule[0.5ex]{1em}{0.4pt}} \textbf{Training} path\\
            \hspace{1.2em} typically collapses\\
            \hspace{1.2em} to driver-blind
        };

    \end{tikzpicture}
    \caption{Arousal–conspecificity framework. The horizontal axis ($\alpha$) controls the balance between internal dynamics (Prior/autocorrelation) and external evidence (Likelihood/drivers). The vertical axis represents conspecificity of driver signals. The desired state is a driver-engaged regime with high arousal and high conspecificity. However, Challenge C1 (architectural shortcuts, blue) pushes models toward low~$\alpha$, while C2 (fidelity gaps, red) reduces perceived driver reliability. Training trajectories (purple dashed) therefore tend to collapse into a driver-blind regime where forecasts rely almost exclusively on autocorrelation.}
\label{fig:arousal_conspecificity}
\end{figure*}
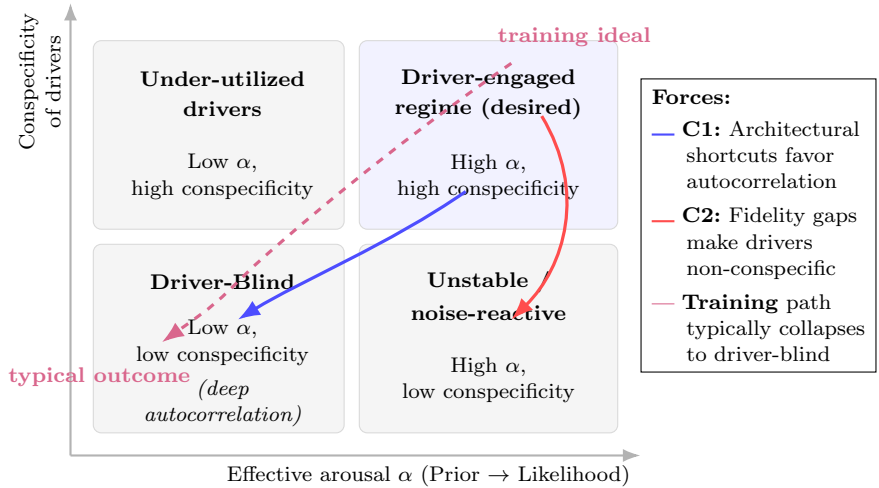

Figure~\ref{fig:arousal_conspecificity} summarizes this view. The horizontal axis represents effective arousal $\alpha$, interpolating from prior-dominated to likelihood-dominated regimes. The vertical axis represents conspecificity of driver signals, from scrambled to well aligned. The desired regime for clinical decision support is high arousal and high conspecificity, where driver events are trusted and strongly influence forecasts. In practice, architectural shortcuts (C1) and data fidelity gaps (C2) push training trajectories toward regimes with low arousal and low perceived conspecificity, so that typical optimization runs collapse into a driver-blind corner of the space. We note that while effective arousal $\alpha$ can in principle be estimated from the relative weighting of loss terms or attention allocated to driver versus CGM channels, conspecificity is more difficult to quantify directly. Operationally, we treat conspecificity as high when driver signals are temporally aligned, smoothly varying, and statistically consistent with outcome changes—conditions approximated by physiological encoders such as IOB/COB curves—and low when drivers appear as sparse, noisy impulses with inconsistent relationships to glucose responses.

\subsection{Challenge C1: Architectural bias and representation limits}

The first source of collapse is architectural or the model's ontological commitment \citep{shakeri2025metaphysics}. Standard deep forecasting architectures provide many ways to model short-range trends and local autocorrelation but relatively few inductive biases that encourage learning long-delayed, heterogeneous driver effects. CGM provides a dense, smooth five-minute signal. Driver channels appear as sparse impulses with magnitudes that vary over orders of magnitude across and within individuals. When such heteroskedastic inputs are concatenated and passed through shared layers, gradient flow and attention weights tend to favor the high-variance, densely sampled channel, i.e.\ glucose.

Temporal encoding machinery also plays a role. Sinusoidal or learned positional encodings assume relatively smooth temporal structure \citep{vaswani2017attention}, but physiologically relevant glucose responses depend on specific lags and convolutions: a bolus at time $t$ affects glucose over several hours with a characteristic impulse response; a meal’s effect depends on gastric emptying and gut absorption; circadian effects modulate insulin sensitivity over the day \cite{egi2007circadian}. Without explicit temporal kernels or architectures designed to represent these convolutions, it is easier for the optimizer to treat driver events as nuisances and rely on extrapolating recent CGM trajectories.

Residual connections and skip pathways compound this by enabling information from the glucose channel to bypass fusion layers \cite{he2016deep}. If a model can achieve competitive loss by projecting CGM through a series of autoregressive or convolutional layers, then the capacity devoted to fusion with drivers may never be fully used. This is particularly true when the training objective is dominated by short horizons where history alone is already extremely predictive. Attention mechanisms can in principle overcome this, but in practice softmax attention over heterogeneous inputs is numerically fragile. When logits differ in scale because of differences in variance or sampling frequency, attention distributions can saturate, leading to effective rank collapse of attention maps \citep{dong2021attention, liang2024lseattention}. 
Recent work on numerically stabilized and channel-aware attention for time-series forecasting mitigates some of these issues by rescaling heterogeneous inputs and preventing softmax saturation \citep{liang2024lseattention, zhu2023edge}. In practice, models that combine such stabilized attention blocks with driver-specific encoders report more balanced attributions across CGM, temporal, and treatment channels, and achieve modest but non-zero $\Delta_{\text{drivers}}$.


Flexible hybrid architectures, such as neural ODEs coupled to mechanistic compartments, do not automatically fix this problem. Instead, they widen the space of functions that can be fitted to confounded observational data. Zou et al.\ \citep{zou2024hybrid} showed that hybrid models trained solely on predictive loss can override monotonicity and sign constraints that physiologists would consider non-negotiable. In our framework, such models inhabit regions of parameter space where internal dynamics are expressive but poorly grounded in the true roles of drivers, a particularly dangerous form of C1.

\subsection{Challenge C2: Fidelity gap and behavioral confounding}

The second source of collapse lies in the data. Real-world logs contain temporal misalignment, magnitude noise, missing events, and unobserved confounders that systematically erode the signal-to-noise ratio of driver channels. Meal times in diaries are often off by fifteen to thirty minutes; carbohydrate estimates can be wrong by twenty to forty percent \citep{brazeau2013carbohydrate}; snacks and corrections are frequently omitted. Activity is reduced to coarse step counts or binary flags that do not distinguish intensity or modality \citep{kemmer1992prevention}. Physiologically important modulators such as stress hormones, sleep quality, acute illness, and menstrual phase are rarely recorded \citep{gonder2016psychology}.

From the model’s point of view, this means that the conditional distribution of future glucose given observed drivers is broad and sometimes multi-modal. Probabilistic architectures such as quantile forecasters and mixture models make this explicit: Zhu et al.\ \citep{zhu2023edge} and Eberle et al.\ \citep{eberle2023blood} used quantile losses to model uncertainty, while Sergazinov et al.\ \citep{sergazinov2023gluformer} employed an infinite mixture model to capture multi-modality. These approaches acknowledge C2, but they also illustrate why deterministic models might rationally choose to downweight drivers. When driver channels appear unreliable, the expected improvement in pointwise loss from using them may be small or negative.

Behavioral confounding exacerbates this. Because patients administer insulin in response to anticipated meals and hyperglycemia, and because many studies train only on observational trajectories, the joint distribution of inputs and outputs under the behavior policy may not contain enough variation to identify treatment effects. In such settings, flexible networks minimize predictive loss by exploiting correlations between glucose and interventions that arise from shared dependence on latent factors such as physician guidance and self-management habits, rather than from the downstream physiological pathways that matter for counterfactual evaluation. In our terminology, fidelity gaps reduce conspecificity: drivers do not look like clean, well-timed inputs to the internal dynamical system but rather like noisy reflections of unmodeled processes. Under those conditions, a low-$\alpha$, driver-blind regime is again a locally optimal behavior for a risk-neutral optimizer.

\subsection{Challenge C3: Personalization gap and non-stationarity}

The third challenge is physiological heterogeneity. Insulin sensitivity varies across individuals by an order of magnitude; insulin-to-carbohydrate ratios can differ by factors of three or four \citep{man2014uva}; basal patterns, dawn phenomenon, and other metabolic traits define distinct response phenotypes \citep{porcellati2013thirty}. Within individuals, insulin sensitivity is non-stationary: it exhibits strong circadian modulation \citep{egi2007circadian}, varies across menstrual cycles \citep{trout2007menstrual}, and is affected by exercise and illness \citep{kemmer1992prevention, gonder2016psychology}.
A fixed set of weights cannot simultaneously represent all these regimes without either very large capacity or mechanisms for rapid on-line adaptation.

If a model is trained on pooled data without adequate personalization, it tends to learn averages that do not match anyone well. Mu\~noz-Organero \citep{munoz2020deep} showed that models trained on one individual generalize poorly to another, with cross-subject RMSE more than four times higher than within-subject RMSE. Personalized architectures with patient embeddings \citep{armandpour2021deep, sergazinov2023gluformer, deng2024pretrained} and multitask setups with patient-specific heads \citep{daniels2022multitask} address C3 by giving the model subject-specific latent context. Nevertheless, when such personalization is absent or insufficient, drivers become even harder to interpret: what looks like a 10-unit insulin bolus may have very different consequences across individuals and across days, further lowering the perceived value of driver channels and reinforcing the tendency toward deep autocorrelation.

Taken together, C1, C2, and C3 explain why many multivariate forecasting models trained on CGM plus drivers converge to behavior where $\Delta_{\text{drivers}}$ is negligible. Architectures make it easy to achieve low loss using only CGM; data make driver channels noisy and confounded; heterogeneity makes driver effects highly context-dependent. In the arousal–conspecificity plane, these pressures push training trajectories away from the driver-engaged regime and into the driver-blind corner.

\section{Discussion: Implications, partial solutions, and recommendations}

The Driver-Blindness phenomenon has several important implications for how the community should design models, evaluate them, and report results. From the perspective of the broader time-series forecasting literature, blood glucose prediction in Type~1 diabetes can be viewed as an adversarial stress test: if generic benchmarks already reveal a tendency for modern architectures to fall back onto linear trends and deep autocorrelation \citep{zeng2023transformers, chencloser}, it is unsurprising that the same biases manifest more sharply in CGM forecasting, where clinically meaningful gains require extracting delayed, dose-dependent effects of insulin and meals from noisy observational logs. It also highlights patterns in existing work that mitigate collapse, along with gaps that remain.

A first implication is that the mere inclusion of driver variables in the input does not guarantee the model has learned to use them. Reported improvements in RMSE or mean absolute error can often be attributed entirely to architectural changes, regularization, or training tricks that also benefit CGM-only baselines. Without an explicit comparison between a multivariate model and a matched univariate model from the same architectural family, using the same training and evaluation protocol, it is impossible to know whether the drivers are contributing meaningfully. This motivates a reporting standard in which $\Delta_{\text{drivers}}$ from \eqref{eq:delta_drivers} is routinely measured at clinically relevant horizons and presented alongside conventional error metrics.

A second implication is that architectures must be co-designed with representations that make drivers conspecific with internal dynamics. The most successful approaches in the literature make driver events look like slowly varying physiological state variables before they are presented to generic deep networks. Bertachi et al.\ \citep{bertachi2018prediction} transform discrete insulin and meal logs into continuous insulin-on-board and carbs-on-board curves (IOB/COB) using fixed pharmacokinetic models; Zecchin et al.\ \citep{zecchin2012neural} feed glucose rate-of-appearance (RaG) signals derived from compartmental models; Mu\~noz-Organero \citep{munoz2020deep} uses separate recurrent branches for insulin, carbohydrate, and glucose subsystems. In all these cases, the fusion problem presented to the neural network is simplified: instead of learning highly heterogeneous, delayed responses from sparse impulses, the network sees smooth, aligned state trajectories that are easier to integrate. Dilated convolutions and carefully chosen dilation schedules \citep{li2020glunet, eberle2023blood} complement this by ensuring that receptive fields cover the relevant lag structure.

A third implication concerns probabilistic forecasting and robustness. Architectures such as E-TFT \citep{zhu2023edge} and the quantile-based TFT variant in \citep{eberle2023blood} acknowledge that uncertainty from C2 cannot be eliminated and therefore predict multiple quantiles rather than single points. Gluformer \citep{sergazinov2023gluformer} goes further by modeling the entire conditional distribution using an infinite mixture model. Robust training strategies, such as minimizing the lower quantile of batch losses \citep{armandpour2021deep}, deliberately ignore the worst outliers that often arise from logging errors. These probabilistic and robust methods do not directly increase $\Delta_{\text{drivers}}$ by themselves, but they do produce models that are better calibrated for risk and more honest about the information content of inputs. They also supply additional diagnostic tools: if the posterior distribution of forecasts is essentially unchanged when drivers are perturbed within realistic bounds, this is another signature of Driver-Blindness.

A fourth implication is that causal structure must be reintroduced, explicitly, into training objectives when working with flexible architectures and confounded observational data. H2NCM \citep{zou2024hybrid} provides a concrete example. By augmenting the predictive loss with a causal ranking loss that penalizes violations of known treatment-ordering constraints, they steer optimization away from spurious regions of parameter space that fit the behavior policy but violate basic pharmacological intuition. Their results indicate that it is possible to maintain state-of-the-art predictive performance while reducing causal error dramatically. More generally, the glucose forecasting community could adopt domain-specific regularizers that enforce monotonicity (for example, more insulin should not increase glucose all else equal), non-negativity of certain impulse responses, or consistency with simple compartmental models. These constraints effectively raise the ``arousal'' parameter $\alpha$ in \eqref{eq:free_energy} for driver channels: they force the model to move beyond deep autocorrelation when doing so is necessary to satisfy causal priors.

Personalization strategies address C3 and have their own implications. Learned patient embeddings \citep{armandpour2021deep, sergazinov2023gluformer, deng2024pretrained} allow shared networks to adapt to individual baselines and sensitivities; multitask recurrent architectures with patient-specific heads \citep{daniels2022multitask} help separate common patterns from idiosyncratic ones. These approaches reduce the mismatch between the population-averaged effect of a driver and its effect on a given individual, increasing conspecificity. However, many studies still report performance on pooled test sets without separating within-subject from cross-subject error, and few explore the speed and stability with which models adapt to distribution shifts induced by changes in therapy, lifestyle, or hormonal state. Incorporating systematic personalization experiments—for example, few-shot fine-tuning curves or held-out subject evaluations—into standard benchmarks would clarify where C3 remains a limiting factor.

Despite these partial successes, important gaps remain. Many promising architectural ideas, including patch-based Transformers for event-level representations \citep{karagoz2025comparative}, multimodal cross-attention designs \citep{isaac2025attengluco}, and state-space models tailored to long-context physiological signals \citep{isaac2025ssmcgm}, have not yet been evaluated under a standardized Driver-Blindness protocol. Even when $\Delta_{\text{drivers}}$ is positive, few studies systematically test robustness under perturbations that mimic realistic logging errors, such as time jitter, missing events, and dose misreporting. Causal validity is almost never directly assessed; exceptions such as \citep{zou2024hybrid} remain rare. There is also a structural bias in widely used benchmarks: when datasets and metrics primarily reward short-horizon error reduction, they implicitly valorize deep autocorrelation models and make it harder to justify the complexity of driver-engaging architectures.

Based on this analysis, several recommendations follow. First, future work on multivariate glucose forecasting should always report $\Delta_{\text{drivers}}$ for matched pairs of univariate and multivariate models at multiple horizons, along with confidence intervals. This simple addition would make Driver-Blindness visible rather than implicit. Second, authors should prefer representations in which drivers are transformed into physiologically meaningful, continuous state variables before being passed to generic deep networks, as in IOB/COB/RaG encoders \citep{bertachi2018prediction, zecchin2012neural}. Third, when using highly flexible models on observational data, training objectives should include causal regularization terms that encode basic domain knowledge, following the spirit of H2NCM. Fourth, evaluation protocols should incorporate robustness and counterfactual tests: for example, isolated insulin-only and carb-only scenarios in simulators such as UVA/Padova \citep{man2014uva}, where the correct direction and timing of responses is known, and stress tests where logging noise is systematically varied. Finally, personalization should be treated as a first-class design objective, with explicit reporting of cross-subject generalization and adaptation behavior.
We acknowledge that several of the proposed mitigations incur additional computational costs relative to simple autoregressive baselines. However, given that glucose forecasting models are typically trained offline and deployed on edge devices or cloud infrastructure with modest inference budgets, we expect training-time overhead to be acceptable in most practical settings. The key constraint is often data availability and quality rather than compute.

Glucose forecasting in Type~1 diabetes is an ideal benchmark for multivariate time series with sparse, delayed exogenous signals because the underlying physiology is relatively well understood, high-fidelity simulators exist, and the stakes for decision support are high. The same design and evaluation principles, however, apply more broadly to multivariate time series with sparse exogenous drivers, from other biomedical time series to climate and macroeconomic forecasting. Addressing Driver-Blindness in this concrete domain can therefore serve as a stepping stone toward a more general science of deep learning for dynamical systems driven by intermittent interventions.

\bibliography{ifacconf}  

\end{document}